\documentclass[conference]{IEEEtran}
\IEEEpubid{}

\IEEEoverridecommandlockouts
\usepackage{cite}
\usepackage{amsmath,amssymb,amsfonts}
\usepackage{graphicx}
\usepackage{textcomp}
\usepackage{physics}
\usepackage{hyperref}
\usepackage{xcolor}
 \usepackage{float}
\usepackage{tabularx,booktabs}
\usepackage{notoccite}
\usepackage{amssymb}
\usepackage[linesnumbered,ruled,vlined]{algorithm2e}
\usepackage[numbers,sort&compress]{natbib}
\renewcommand{\footnoterule}{%
  \kern -3pt
  \hrule width 2in
  \kern 2pt
}

\SetCommentSty{mycommfont}

\SetKwInput{KwInput}{Input}                
\SetKwInput{KwOutput}{Output} 

\def\BibTeX{{\rm B\kern-.05em{\sc i\kern-.025em b}\kern-.08em
    T\kern-.1667em\lower.7ex\hbox{E}\kern-.125emX}}
\newcommand{\sample}{0.5\linewidth}

\makeatletter
\newcommand*{\rom}[1]{\expandafter\@slowromancap\romannumeral #1@}
\makeatother
\begin{document}

\title{Lipschitz Bound Analysis of Neural Networks\\}

\author{\IEEEauthorblockN{Sarosij Bose}
\IEEEauthorblockA{\textit{Department of Computer Science and Engineering} \\
\textit{University of Calcutta}\\
Kolkata, India \\
sarosijbose2000@gmail.com}
\thanks{Pre-print. Under Review}
}






\maketitle

\begin{abstract}
Lipschitz Bound Estimation is an effective method of regularizing deep neural networks to make them robust against adversarial attacks. This is useful in a variety of applications ranging from reinforcement learning to autonomous systems. In this paper, we highlight the significant gap in obtaining a non-trivial Lipschitz bound certificate for Convolutional Neural Networks (CNNs) and empirically support it with extensive graphical analysis. We also show that unrolling Convolutional layers or Toeplitz matrices can be employed to convert Convolutional Neural Networks (CNNs) to a Fully Connected Network. Further, we propose a simple algorithm to show the existing 20x-50x gap in a particular data distribution between the actual lipschitz constant and the obtained tight bound. We also ran sets of thorough experiments on various network architectures and benchmark them on datasets like MNIST and CIFAR-10. All these proposals are supported by extensive testing, graphs, histograms and comparative analysis.
\end{abstract}
\bigskip

\section{Introduction}
\par Adversarial attacks \cite{goodfellow2014explaining} on neural networks was first demonstrated by Szegedy et. al. which showed that a fully connected neural network could be made to falsely classify MNIST \cite{lecun1998mnist} images. This has raised questions over the reliability of such networks when used in critical real time applications such as self-driving cars \cite{kumar2020black}. A host of regularizing techniques have been proposed thereof for tackling this problem such as batch normalization \cite{ioffe2015batch}, weight decaying \cite{loshchilov2017decoupled} etc. Accordingly, using the Lipschitz constant as a certificate for robustness of models has been studied extensively in the past in several works such as in \cite{weng2018evaluating}, \cite{wong2018scaling} and \cite{ruan2018reachability}.
\par In this paper, we consider the problem of using the Lipschitz constant as a regularizer for neural networks.
The lipschitz constant for any function $f: \mathbb{R}\textsuperscript{n} \rightarrow \mathbb{R}\textsuperscript{m}$ is defined as an non-negative integer $L > 0$ such that:

\begin{equation}
\label{eq:formula1}
 ||f(x) - f(y)|| < L ||x - y||
\end{equation}

The smallest value of L which satisfies the criterion in \ref{eq:formula1} is known as the Lipschitz constant for $f$. This function $f$ is characterized by a neural network in this work and $x$ and $y$ can be pairs of different images. Framing the equation in this setting is extremely useful for a host of applications such as Feedback loops in control theory \cite{aswani2013provably} or to train safe agents in Reinforcement Learning \cite{berkenkamp2017safe}.
\par In this work, we are mainly concerned about two approaches to calculate the lipschitz constant. In \cite{szegedy2013intriguing}, the authors obtain the global lipschitz constant of a fully connected neural network by taking the product of the lipschitz constants of the individual layers. In \cite{fazlyab2019efficient}, Fazlyab et al. introduces a novel technique where the non-linear activation functions present in any Fully Connected Neural Network could be projected into an operator ball and regulated to obtain a tight upper Lipschitz bound. However, this approach suffered from some drawbacks such as expensive semi-definite programming which is not scalable. Further, this approach is limited to only Fully Connected Networks and not directly applicable to other deep neural networks such as CNNs.
\par In \cite{sedghi2018singular}, researchers from Google used the singular values of each convolutional layer separately. Taking the product of the maximum from the range of these singular values produced could hence yield the trivial Lipchitz bound of the CNN. They showed a fast approach where there is no need to store the kernel matrices in memory. Instead, just the filter itself and the input image dimensions ($m \times n \times c$)\footnote{m = height of the image, n = width of the image and c = number of input channels.} per layer would suffice. However, it was unable to find out tight Lipschitz bounds since it does not take into account any non-linearity such as activation functions which may be present in the network.
\par We first discuss the methods by which Convolutional Neural Networks can be converted to fully connected networks such as unrolling the kernels present in each layer \cite{chellapilla2006high} or employing toeplitz matrices \cite{araujo2021lipschitz}. With these converted CNNs, we report our observations and results on the various bounds obtained. Next, through our extensive histogram illustrations, we show the distribution of lipschitz constants over the MNIST and CIFAR-10 \cite{krizhevsky2009learning} datasets. This directly gives us the exact range of bounds we can ideally expect from a perfectly robust network. We do this for both fully connected networks and CNNs separately. 
\par Finally, we propose a simple algorithm to compare the trivial bound, tight bound and the average of the maximum empirical bounds for a given particular fully connected network and graphically analyze their behaviour over extended batches of images. 

Our contributions in this work can be briefly summarised as follows:
\begin{itemize}

\item We conduct an extensive study of the various methods used for finding the lipschitz constant for Fully Connected Networks and CNNs.
\item Through our extensive experiments, we empirically show that the trivial bound is still significantly higher than the actual lipschitz constant.
\item We show how fully connected networks can be converted to convolutional neural networks and it's tight lipschitz bound found out.
\item We propose a simple algorithm to show how the averaged value of lipschitz constant slowly approaches the maximum lipschitz value for each batch of images.

\end{itemize}

The rest of the paper is organised in the following manner: In Section II we discuss the conversion of CNNs to fully connected networks, in Section III we analyze the various lipschitz bounds, in Section IV we detail our experimental setup and finally in Section V we present the conclusion.

\section{Conversion of CNNs to FCNs}

\subsection{Toeplitz Matrices}

Toeplitz matrices are a special subset of doubly block circulant matrices. The filters of each individual convolutional layer are re-defined to form a sparse matrix which is then multiplied with the reshaped input image vector. Hence, convolution can be implemented as a matrix-vector product with the help of such matrices. This is particularly necessary for the CNN to satisfy equation \ref{eq:formula2} and hence behave as a fully connected network. This approach has been attempted before such as in \cite{araujo2021lipschitz}, where the authors try to estimate a tight lipschitz bound from convolutional layers.

\subsection{Unrolling Convolution}

The concept of unrolling convolution was first proposed in \cite{chellapilla2006high} in 2006 and improved in \cite{vasudevan2017parallel}. The authors proposed to convert the process of convolution into a matrix only based approach. Instead of individually performing convolution over separate input features with each filter, they propose to unfold the input images and stitch together the kernels for each layer to form a single unified input feature matrix and a large weight matrix respectively. The multiplication between these newly formed feature matrix and weight matrix produces the “unrolled” output of convolution which on rolling back produces the same output as traditional convolution. The step involving the creation of the feature matrix and weight matrix is called as “unrolling”. The primary motivation this is to mimic the properties of a $l$ layered fully connected network which can be expressed by the equation below:

\begin{equation}
\label{eq:formula2}
 f(x) = W\textsuperscript{l}x\textsuperscript{l} + b\textsuperscript{l}
\end{equation}

\par Where $x \in R$ is an initial input to the network, $W \in R\textsuperscript{2}$ is the weight matrix and $b \in R$ is the bias vector. By “unrolling” the convolution operation, it can be successfully converted into the form shown in Fig. 1.\par
\begin{figure}[ht]
		\centering
		\includegraphics[height=0.6\linewidth]{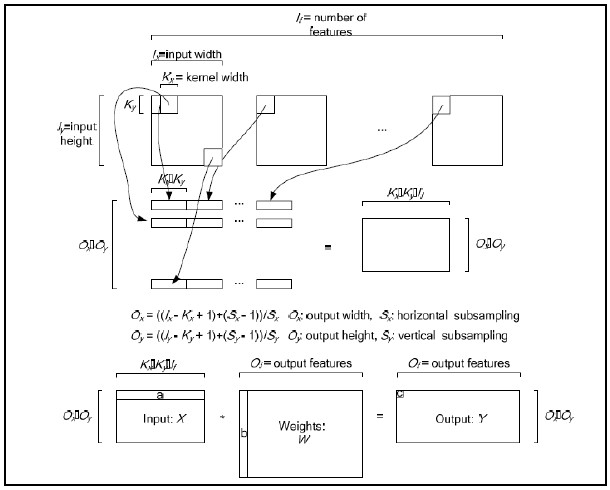}
		\caption{\label{fig:unroll}The process of “unrolling” convolution from \cite{chellapilla2006high}. (Biases, sub-sampling, and non-linearity omitted from here as the focus is only on the conversion of the convolution into a matrix product)}
	\end{figure}
In the given image, K\textsubscript X and K\textsubscript Y represent the dimension of the kernels. (I\textsubscript X , I\textsubscript Y) represents the dimensions of the input image, (O\textsubscript X , O\textsubscript Y) represents the dimensions of the output matrix and (S\textsubscript X , S\textsubscript Y) represents the sub-sampling in the respective directions. \cite{bose2021rescnn}

\section{Lipschitz Bound Analysis}
\subsection{Empirical Bound Distribution}
\par Using formula \ref{eq:formula1}, given an encoder $f(x)$ and when two distinct images are placed in place of $x$ and $y$ are repeated over various shuffled pairs of images, the Histogram shown in Figure 2a is obtained. The corresponding trivial and tight lipschitz bounds are 2041.604 and 800.502 respectively. However, the maximum value of the empirical bound in the histogram is 18.91. Hence, there is a 107x and 40x difference of the trivial and tight bounds over the empirical bound respectively. We can also see how the actual lipschitz constant behaves over a particular distribution. All input images were taken from the test split of the MNIST dataset.

\renewcommand{\thefigure}{2a}
\begin{figure}[ht]
		\centering
		\includegraphics[height=\sample, width=\linewidth]{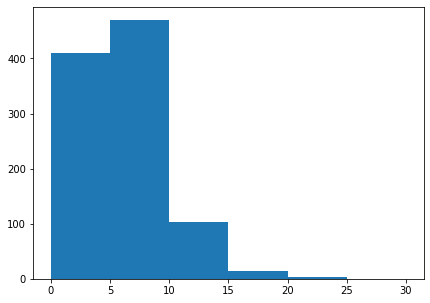}
		\caption{\label{fig:hfc} The histogram of empirical bounds for Fully Connected Network shown in Fig 7. The x-axis corresponds to the values of bounds obtained and the y-axis shows the frequency of the bounds within each bracket.}
	\end{figure}
\par Similarly, following the same procedure outlined above for a CNN, we obtain the histogram shown below in Figure 2b. Here, the trivial lipschitz bound obtained is 733.248 which is again 325x higher than the obtained emipirical bound of only 2.25. All input images were taken from the test split of the CIFAR-10 dataset.
\renewcommand{\thefigure}{2b}
\begin{figure}[h]
		\centering
		\includegraphics[height=\sample]{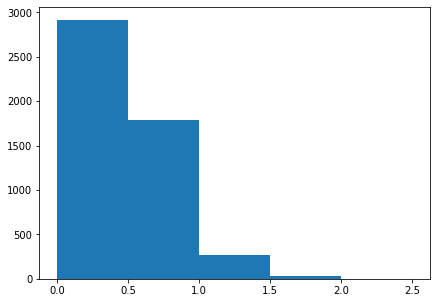}
		\caption{\label{fig:hcnn} The histogram of empirical bounds for CNN shown in Fig 8. The x-axis corresponds to the values of bounds obtained and the y-axis shows the frequency of the bounds within each bracket.}
	\end{figure}

A rigorous empirical analysis is important since the lipschitz constant for a particular deep network can't be exactly determined \cite{virmaux2018lipschitz}. From the above histograms, a couple of interesting observations can be made. The actual lipschitz bound for a CNN is much lower than a similar sized Fully Connected Network which shows how tight a comparable CNN lipschitz regularizer needs to be to provide a robustness certificate. The Deep Networks used here, the corresponding datasets and other metrics can be found in detail in Section \rom{4}.

\subsection{Trivial Bounds}\label{AA}
\par The trivial lipschitz bound for a particular feed forward neural network refers to the product of the spectral norms of the weight matrix of each individual layer. For example, for a given l layered network $f(x)$ which has two consecutive layers \{k, k+1\} with $n1$ and $n2$ number of neurons respectively, the corresponding weight matrix is $W \in R\textsuperscript{n1 x n2}$. Accordingly, the spectral norm for each layer k is $\norm{W}\textsubscript{2}$. The trivial lipschitz bound L\textsubscript{p} can thus be given by:-
\begin{equation}
\label{eq:formula3}
 L\textsubscript{p} = \prod_{k=1}^{l} \norm{W}\textsubscript{2}
\end{equation}

\par As highlighted in Section \rom{2}, for CNNs the individual filters present in each convolutional layer can be either "unrolled" to form a single combined representative filter or converted to toeplitz matrices to satify formula \ref{eq:formula1}. The trivial lipschitz bound for the CNN can then be found out using \ref{eq:formula3}.

\subsection{Convergence of Bounds}
We graphically illustrate here how the trivial, tight and empirical lipschitz bound varies over a particular data distribution. Hence, we propose Algorithm 1 below to observe how the three lipschitz bounds behave over the MNIST and CIFAR-10 datasets. The individual empirical lipschitz values for each set of images are listed out in Table \rom{1} and \rom{2}.

\begin{algorithm}
\DontPrintSemicolon
  
  \KwInput{Set Size N, Pre-trained model $f(x)$}
  \KwOutput{List of Max. Empirical Lipschitz bounds L\textsubscript{emp}}
  \KwData{Test Split of Dataset D}
  L = [] \\
  itermax = 0\\
  batches $\leftarrow$ Shuffle(D, N)\\
  \For{batch in batches}
    {
        $pairs = {batch \choose 2}$\\
        \For{pair in pairs}
            {
                image1 $\leftarrow$ pair[0]\\
                output1 $\leftarrow$ f(image1)\\
                image2 $\leftarrow$ pair[1]\\
                output2 $\leftarrow$ f(image2)\\
                L\textsubscript{emp} = $\dfrac{\norm{output2-output1}\textsubscript{2}} {\norm{image2-image1}\textsubscript{2}}$\\
                \If{$L\textsubscript{emp} > itermax$}
                    {
                        itermax = L\textsubscript{emp} 
                    }
            }
        Append itermax to L\\
    }
  Get L, itermax, L\textsubscript{emp}

\caption{Empirical lipschitz constant over a data distribution}
\end{algorithm}

\renewcommand{\thefigure}{2c}
\begin{figure}
	\centering
	\frame{\includegraphics[height=\sample]{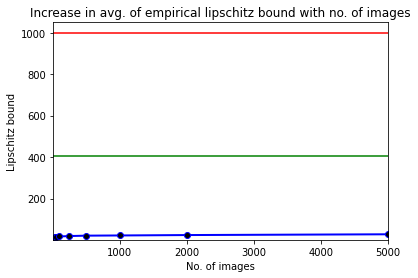}}
	\caption{\label{fig:bounds}The red line denotes the trivial lipschitz bound, the green line denotes the tight bound and the blue line denotes the actual lipschitz constants.}
\end{figure}

The value of the trivial bound is 997.54 and the tight bound is 406.720. In comparison, the maximum empirical lipschitz constant is only 27.93. Data for different batches can be found in Table \rom{1} for the MNIST Dataset. 

\begin{center}
\begin{table}[h]
\centering
\caption{AVERAGE AND MAXIMUM EMPIRICAL LIPSCHITZ BOUNDS OBTAINED ON THE MNIST DATASET}
\begin{tabular}{||c c c||} 
 \hline
 Set Size (N) & Avg. emp value & Max. of max emp value \\ [0.5ex] 
 \hline\hline
 50 & 15.24 & 20.64 \\ 
 \hline
 250 & 19.07 & 24.76 \\ 
 \hline
 500 & 21.00 & 23.03 \\ 
 \hline
 1000 & 21.92 & 24.37 \\ 
 \hline
 2000 & 23.88 & 27.11 \\ 
 \hline
5000 & 27.59 & 27.93 \\ [1ex] 
 \hline
\end{tabular}
\end{table}
\end{center}

\renewcommand{\thefigure}{2d}
\begin{figure}
		\centering
		\frame{\includegraphics[height=\sample]{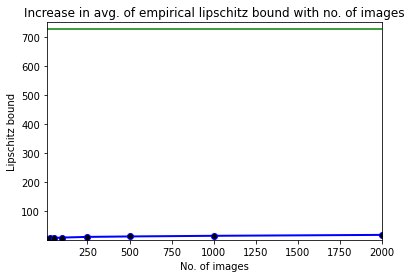}}
		\caption{\label{fig:comp} The green line denotes the trivial bound and the blue line denotes the empirical lipschitz constants. The tight bound could'nt be reported here due to resource constraints.}
	\end{figure}
\par The trivial lipschitz bound in this case is 727.091 in this case which as evident from figure 2d is far higher than the maximum actual lipschitz constant which is 19.28. In Table \rom{2}, we report the values obtained from the same procedure for the CIFAR-10 dataset.  
\begin{center}
\begin{table}[h]
\centering
\caption{AVERAGE AND MAXIMUM EMPIRICAL LIPSCHITZ BOUNDS OBTAINED ON THE CIFAR-10 DATASET}
\begin{tabular}{||c c c||} 
 \hline
 Set Size (N) & Avg. emp value & Max. of max emp value \\ [0.5ex] 
 \hline\hline
 50 & 7.12 & 14.99 \\ 
 \hline
 100 & 8.71 & 17.31 \\ 
 \hline
 250 & 11.00 & 15.88 \\ 
 \hline
 500 & 12.54 & 19.28 \\ 
 \hline
 1000 & 14.76 & 19.25 \\ 
 \hline
2000 & 17.90 & 19.28 \\ [1ex] 
 \hline
\end{tabular}
\end{table}
\end{center}

\par From the last 2 rows of Table \rom{1} and \rom{2}, it is clearly evident how the average empirical lipschitz value and the maximum slowly converge and saturate towards a particular fixed constant. 
 
\section{Experiments}
\textbf{Implementation Details}: Given below is the training and testing regimen followed for the Fully Connected Networks and Convolutional Neural Networks. 

\begin{itemize}

\item We train all our models on the MNIST and CIFAR-10 datasets. The MNIST Dataset of $28 \times 28$ greyscale digit images consisting of 10 classes. There are a total of 60, 000 training images and 10, 000 for validation/testing purposes. The CIFAR-10 Dataset consists of 10 classes of $32 \times 32$ RGB images of common animals. There are a total of 50, 000 training images and 10, 000 for validation/testing purposes. For obtaining the bounds, only the test splits were used for model inference purposes.
\item The basic architecture for the Fully Connected Network and the CNN we used has been shown in Figure 3a and 3b respectively. Several manual adjustments have been made to the hyperparameters in these networks accordingly to optimize performance and fit our memory requirements.
\item We train the fully connected network described in Figure 3a for 4 epochs. We used the Negative Log Likelihood (NLL) loss function and the Adam optimizer keeping a fixed learning rate of 1e-3. The optimum Batch Size was found to be 128 for all training purposes.
\item We noted that 4 epochs were enough for the fully connected network to achieve 96.9\% test accuracy on the MNIST Dataset. We obtained a validation accuracy of 54\% on the CIFAR-10 Dataset after the CNN was converted to a Fully Connected Network.
\end{itemize}

\renewcommand{\thefigure}{3a}
\begin{figure}[h]
		\centering
		\frame{\includegraphics[height=0.6\linewidth, width=\linewidth]{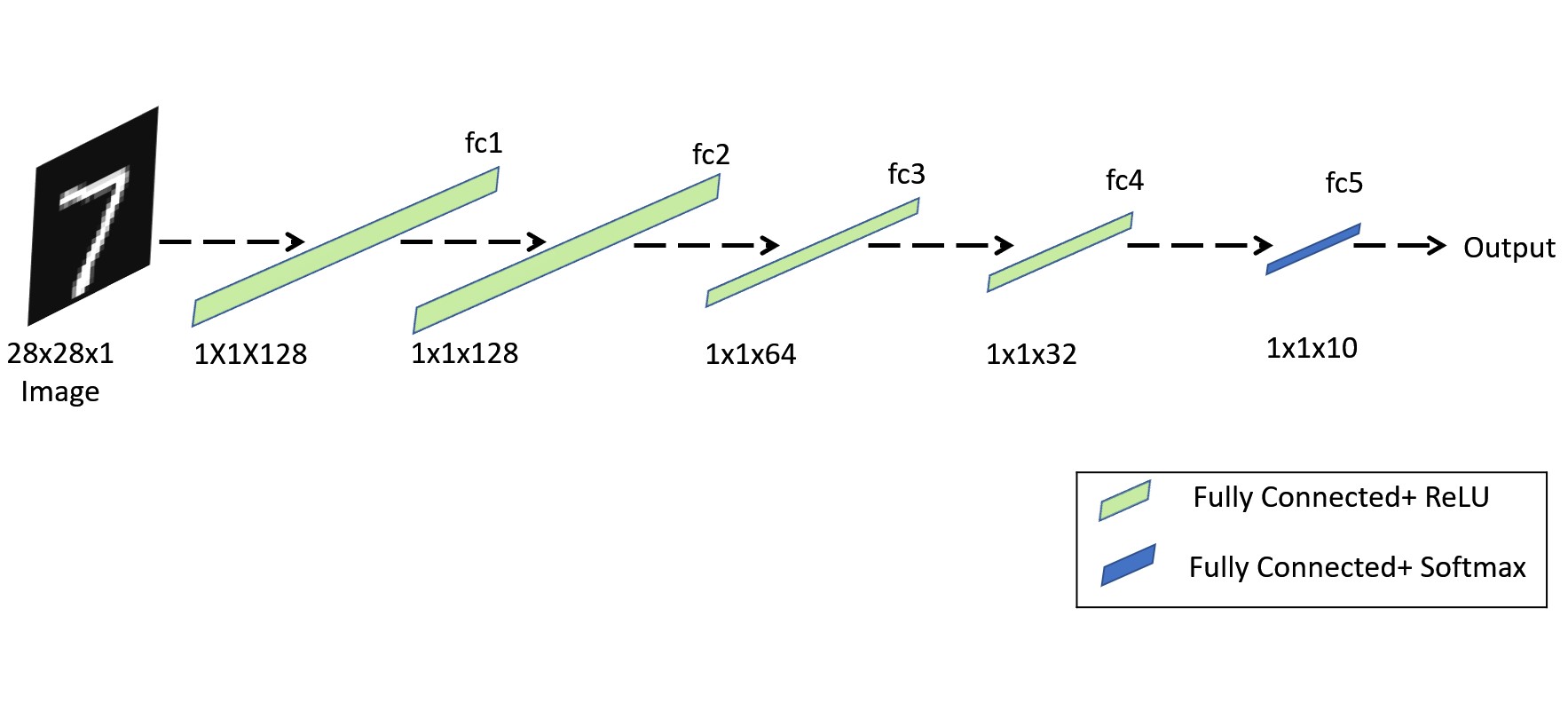}}
		\caption{\label{fig:schema} Schematic Architecture of the Fully Connected Network. The output signifies the predicted class.}
	\end{figure}
\renewcommand{\thefigure}{3b}	
\begin{figure}[h]
		\centering
		\frame{\includegraphics[height=0.6\linewidth, width=\linewidth]{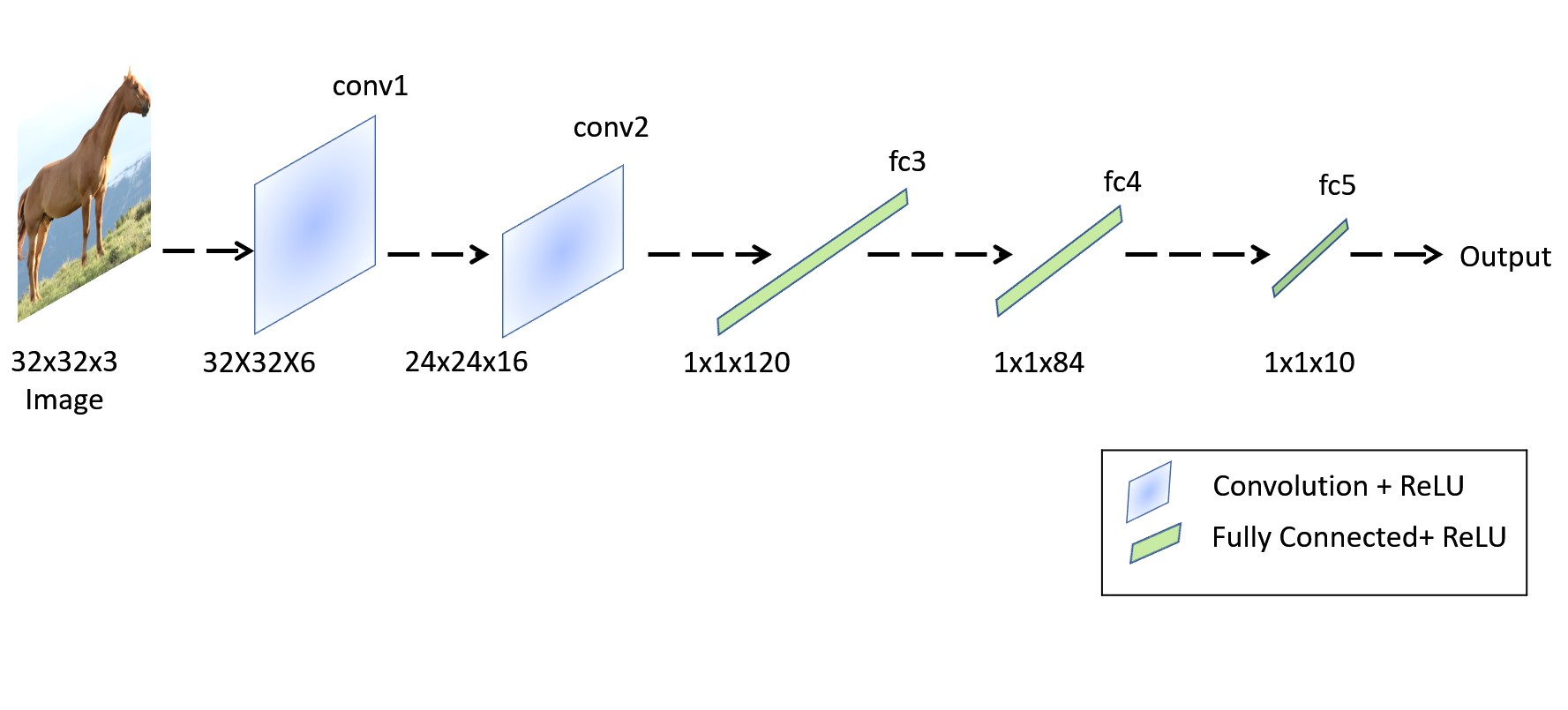}}
		\caption{\label{fig:arch} Schematic Architecture of the Convolutional Neural Network. The output signifies the predicted class.}
	\end{figure}

\section{Conclusion}
\par \par We bring out the strengths and weaknesses of the various existing approaches to apply lipschitz regularization to both Fully Connected Networks and CNNs  and it's numerous applications where it is extensively used like in Autonomous Driving, Control Theory, Reinforcement Learning and GANs. We perform an extensive empirical analysis of the trivial, tight and empirical lipschitz bounds over the MNIST and CIFAR-10 datasets through numerous Fully Connected Networks and CNNs. We illustrate our results through histograms and graphs showing how the empirical lipschitz constant varies over a particular data distribution ($\therefore$ MNIST and CIFAR-10 datasets in our case). We also propose a simple algorithm to show how the average and maximum empirical bounds slowly converge towards the actual bound over each batch iteration and observe how they tend to saturate much below both the trivial and tight bounds. Inspite of these advances, there is still the need to use computationally expensive techniques such as Semi Definite Programming to get a non-trivial bound or use singular values of individual layers to get a trivial bound, hence sacrificing the robustness of the CNN. Further, even the tight bounds are of the order of 20x to 50x higher than the actual lipschitz bound. Future work can be pursued in this direction to eliminate these weaknesses.

\section{Acknowledgement}
I would like to thank my internship advisor Prof. Kunal Narayan Chaudhury for supporting and guiding me throughout the duration of the project and providing the necessary resources for carrying out this work.

\bibliography{refs} 

\begin{thebibliography}{19}
\providecommand{\natexlab}[1]{#1}
\providecommand{\url}[1]{\texttt{#1}}
\expandafter\ifx\csname urlstyle\endcsname\relax
  \providecommand{\doi}[1]{doi: #1}\else
  \providecommand{\doi}{doi: \begingroup \urlstyle{rm}\Url}\fi

\bibitem[Goodfellow et~al.(2014)Goodfellow, Shlens, and
  Szegedy]{goodfellow2014explaining}
Ian~J Goodfellow, Jonathon Shlens, and Christian Szegedy.
\newblock Explaining and harnessing adversarial examples.
\newblock \emph{arXiv preprint arXiv:1412.6572}, 2014.

\bibitem[LeCun(1998)]{lecun1998mnist}
Yann LeCun.
\newblock The mnist database of handwritten digits.
\newblock \emph{http://yann. lecun. com/exdb/mnist/}, 1998.

\bibitem[Kumar et~al.(2020)Kumar, Vishnu, Mitra, and Mohan]{kumar2020black}
K~Naveen Kumar, C~Vishnu, Reshmi Mitra, and C~Krishna Mohan.
\newblock Black-box adversarial attacks in autonomous vehicle technology.
\newblock In \emph{2020 IEEE Applied Imagery Pattern Recognition Workshop
  (AIPR)}, pages 1--7. IEEE, 2020.

\bibitem[Ioffe and Szegedy(2015)]{ioffe2015batch}
Sergey Ioffe and Christian Szegedy.
\newblock Batch normalization: Accelerating deep network training by reducing
  internal covariate shift.
\newblock In \emph{International conference on machine learning}, pages
  448--456. PMLR, 2015.

\bibitem[Loshchilov and Hutter(2017)]{loshchilov2017decoupled}
Ilya Loshchilov and Frank Hutter.
\newblock Decoupled weight decay regularization.
\newblock \emph{arXiv preprint arXiv:1711.05101}, 2017.

\bibitem[Weng et~al.(2018)Weng, Zhang, Chen, Yi, Su, Gao, Hsieh, and
  Daniel]{weng2018evaluating}
Tsui-Wei Weng, Huan Zhang, Pin-Yu Chen, Jinfeng Yi, Dong Su, Yupeng Gao,
  Cho-Jui Hsieh, and Luca Daniel.
\newblock Evaluating the robustness of neural networks: An extreme value theory
  approach.
\newblock \emph{arXiv preprint arXiv:1801.10578}, 2018.

\bibitem[Wong et~al.(2018)Wong, Schmidt, Metzen, and Kolter]{wong2018scaling}
Eric Wong, Frank Schmidt, Jan~Hendrik Metzen, and J~Zico Kolter.
\newblock Scaling provable adversarial defenses.
\newblock \emph{Advances in Neural Information Processing Systems}, 31, 2018.

\bibitem[Ruan et~al.(2018)Ruan, Huang, and Kwiatkowska]{ruan2018reachability}
Wenjie Ruan, Xiaowei Huang, and Marta Kwiatkowska.
\newblock Reachability analysis of deep neural networks with provable
  guarantees.
\newblock \emph{arXiv preprint arXiv:1805.02242}, 2018.

\bibitem[Aswani et~al.(2013)Aswani, Gonzalez, Sastry, and
  Tomlin]{aswani2013provably}
Anil Aswani, Humberto Gonzalez, S~Shankar Sastry, and Claire Tomlin.
\newblock Provably safe and robust learning-based model predictive control.
\newblock \emph{Automatica}, 49\penalty0 (5):\penalty0 1216--1226, 2013.

\bibitem[Berkenkamp et~al.(2017)Berkenkamp, Turchetta, Schoellig, and
  Krause]{berkenkamp2017safe}
Felix Berkenkamp, Matteo Turchetta, Angela Schoellig, and Andreas Krause.
\newblock Safe model-based reinforcement learning with stability guarantees.
\newblock \emph{Advances in neural information processing systems}, 30, 2017.

\bibitem[Szegedy et~al.(2013)Szegedy, Zaremba, Sutskever, Bruna, Erhan,
  Goodfellow, and Fergus]{szegedy2013intriguing}
Christian Szegedy, Wojciech Zaremba, Ilya Sutskever, Joan Bruna, Dumitru Erhan,
  Ian Goodfellow, and Rob Fergus.
\newblock Intriguing properties of neural networks.
\newblock \emph{arXiv preprint arXiv:1312.6199}, 2013.

\bibitem[Fazlyab et~al.(2019)Fazlyab, Robey, Hassani, Morari, and
  Pappas]{fazlyab2019efficient}
Mahyar Fazlyab, Alexander Robey, Hamed Hassani, Manfred Morari, and George
  Pappas.
\newblock Efficient and accurate estimation of lipschitz constants for deep
  neural networks.
\newblock \emph{Advances in Neural Information Processing Systems}, 32, 2019.

\bibitem[Sedghi et~al.(2018)Sedghi, Gupta, and Long]{sedghi2018singular}
Hanie Sedghi, Vineet Gupta, and Philip~M Long.
\newblock The singular values of convolutional layers.
\newblock \emph{arXiv preprint arXiv:1805.10408}, 2018.

\bibitem[Chellapilla et~al.(2006)Chellapilla, Puri, and
  Simard]{chellapilla2006high}
Kumar Chellapilla, Sidd Puri, and Patrice Simard.
\newblock High performance convolutional neural networks for document
  processing.
\newblock In \emph{Tenth international workshop on frontiers in handwriting
  recognition}. Suvisoft, 2006.

\bibitem[Araujo et~al.(2021)Araujo, Negrevergne, Chevaleyre, and
  Atif]{araujo2021lipschitz}
Alexandre Araujo, Benjamin Negrevergne, Yann Chevaleyre, and Jamal Atif.
\newblock On lipschitz regularization of convolutional layers using toeplitz
  matrix theory.
\newblock In \emph{35th AAAI Conference on Artificial Intelligence, Vancouver,
  Canada}, 2021.

\bibitem[Krizhevsky et~al.(2009)Krizhevsky, Hinton,
  et~al.]{krizhevsky2009learning}
Alex Krizhevsky, Geoffrey Hinton, et~al.
\newblock Learning multiple layers of features from tiny images.
\newblock 2009.

\bibitem[Vasudevan et~al.(2017)Vasudevan, Anderson, and
  Gregg]{vasudevan2017parallel}
Aravind Vasudevan, Andrew Anderson, and David Gregg.
\newblock Parallel multi channel convolution using general matrix
  multiplication.
\newblock In \emph{2017 IEEE 28th international conference on
  application-specific systems, architectures and processors (ASAP)}, pages
  19--24. IEEE, 2017.

\bibitem[Bose and Dey(2021)]{bose2021rescnn}
Sarosij Bose and Avirup Dey.
\newblock Rescnn: An alternative implementation of convolutional neural
  networks.
\newblock In \emph{2021 IEEE 8th Uttar Pradesh Section International Conference
  on Electrical, Electronics and Computer Engineering (UPCON)}, pages 1--5.
  IEEE, 2021.

\bibitem[Virmaux and Scaman(2018)]{virmaux2018lipschitz}
Aladin Virmaux and Kevin Scaman.
\newblock Lipschitz regularity of deep neural networks: analysis and efficient
  estimation.
\newblock \emph{Advances in Neural Information Processing Systems}, 31, 2018.

\end{thebibliography}

\end{document}